\documentclass[letterpaper]{article} % DO NOT CHANGE THIS
\usepackage{aaai2026}  % DO NOT CHANGE THIS
\usepackage{times}  % DO NOT CHANGE THIS
\usepackage{helvet}  % DO NOT CHANGE THIS
\usepackage{courier}  % DO NOT CHANGE THIS
\usepackage[hyphens]{url}  % DO NOT CHANGE THIS
\usepackage{graphicx} % DO NOT CHANGE THIS
\usepackage{natbib}  % DO NOT CHANGE THIS AND DO NOT ADD ANY OPTIONS TO IT
\usepackage{caption} % DO NOT CHANGE THIS AND DO NOT ADD ANY OPTIONS TO IT
\usepackage{algorithm}
\usepackage{algorithmic}
\usepackage{amssymb}
\usepackage{graphicx}
\usepackage{amsmath}
\usepackage{multirow}
\usepackage{subcaption}
\usepackage{booktabs}
\usepackage{rotating}

\usepackage{newfloat}
\usepackage{listings}
\DeclareCaptionStyle{ruled}{labelfont=normalfont,labelsep=colon,strut=off} % DO NOT CHANGE THIS
\floatstyle{ruled}
\newfloat{listing}{tb}{lst}{}
\floatname{listing}{Listing}
\title{FDP: A Frequency-Decomposition Preprocessing Pipeline for \\
Unsupervised Anomaly Detection in Brain MRI}
\author{
    Hao Li$^{1}$\equalcontrib,
    Zhenfeng Zhuang$^{1}$\equalcontrib,
    Jingyu Lin$^{1}$\equalcontrib,
    Yu Liu$^{1}$,
    Yifei Chen$^{2}$, \\
    Qiong Peng$^{1}$, 
    Lequan Yu$^{3}$\thanks{Corresponding authors.}, 
    Liansheng Wang$^{1}$\footnotemark[2]
}
\affiliations{
    \textsuperscript{\rm 1}Xiamen University \\
    \textsuperscript{\rm 2}Tencent \\
    \textsuperscript{\rm 3}The University of Hong Kong
}

\usepackage{bibentry}
\begin{document}

% \showauthors
\maketitle

\begin{abstract}
Due to the diversity of brain anatomy and the scarcity of annotated data, supervised anomaly detection for brain MRI remains challenging, driving the development of unsupervised anomaly detection (UAD) approaches. 
Current UAD methods typically utilize artificially generated noise perturbations on healthy MRIs to train generative models for normal anatomy reconstruction, enabling anomaly detection via residual maps.
However, such simulated anomalies lack the biophysical fidelity and morphological complexity characteristic of true clinical lesions.
To advance UAD in brain MRI, we conduct the first systematic frequency-domain analysis of pathological signatures, revealing two key properties: (1) anomalies exhibit unique frequency patterns distinguishable from normal anatomy, and (2) low-frequency signals maintain consistent representations across healthy scans. These insights motivate our Frequency-Decomposition Preprocessing (FDP) framework, the first UAD method to leverage frequency-domain reconstruction for simultaneous pathology suppression and anatomical preservation. FDP can integrate seamlessly with existing anomaly simulation techniques, consistently enhancing detection performance across diverse architectures while maintaining diagnostic fidelity.
Experimental results demonstrate that FDP consistently improves anomaly detection performance when integrated with existing methods. Notably, FDP achieves a $17.63\%$ increase in DICE score with LDM while maintaining robust improvements across multiple baselines. 
% ($\geq$$6.16\%$ DICE, $\geq$$11.37\%$ AUPRC).
\begin{links}
\link{Code}{https://github.com/ls1rius/MRI_FDP}.
\end{links}
\end{abstract}

\section{Introduction}
\label{sec:intro}
Magnetic resonance imaging (MRI) play a crucial role in medical imagine system to provide detailed tissue information without requiring invasive procedures or exposure to radiation for aiding radiologists with their diagnostic and decision-making processes. 
In typical clinical MRI systems, a strong static magnetic field aligns hydrogen nuclei in the body. The nuclei are then excited by radiofrequency (RF) pulses, and as they return to equilibrium they emit detectable signals. Spatial localization is achieved using gradient fields that modulate the phase and frequency of the signal. The measurements are acquired as samples in k-space, a spatial-frequency domain in which each point encodes a spatial-frequency component of the image~\cite{sarracanie2015low, mcrobbie2017mri, rinck2018magnetic}. The spatial-domain MRI is then reconstructed by applying an inverse Fourier transform to the k-space data.

Unsupervised anomaly detection (UAD) methods overcome the limitations of supervised approaches by eliminating the need for annotated pathological data. Typically, these methods learn representations of healthy anatomy through generative models trained on normal MRI scans with artificial noise. During inference, the models process input scans (which may contain anomalies) and attempt their reconstruction under the learned healthy representation. The pixel-wise residuals between original and reconstructed scans serve as anomaly scores, enabling detection of pathologies such as tumors or lesions without requiring abnormal training samples. This paradigm addresses key challenges including annotation scarcity, privacy constraints, and class imbalance, while being applicable to real-world clinical scenarios where abnormalities are rare and exhibit high variability~\cite{bercea2025evaluating, bao2024bmad, behrendt2024patched, kascenas2022denoising, wyatt2022anoddpm}. However, such synthetic noise is limited in reflecting the complexity of real pathological variations, which restricts generalization to clinical cases.

Given the inherent frequency-domain nature of MRI generation (via k-space acquisition), recent studies have increasingly explored frequency-based approaches for reconstruction and analysis~\cite{yi2023frequency, liu2024highly, zou2025mmr}. However, UAD in brain MRI remains predominantly spatial-domain focused, neglecting the diagnostic potential of frequency-component analysis. This represents a significant oversight, as k-space data naturally encodes structural information through distinct frequency components - features that conventional image-space methods cannot directly access. The analytical potential of frequency-domain representations remains particularly underutilized for characterizing pathological deviations in unsupervised settings.

In this paper, we conduct the first systematic frequency-domain analysis of brain MRI anomalies, focusing on: (1) characterizing the intrinsic relationship between pathological features and their frequency signatures, (2) verifying the consistency of the low frequency representations, and (3) developing Frequency-Decomposition Preprocessing (FDP) - a modular component that mitigates pathological artifacts by low-frequency reconstruction while preserving diagnostically critical high-frequency details. 
This module could directly enhances existing anomaly simulation methods through its capability to separate and process pathological signatures in the frequency domain, significantly improving the quality of synthesized healthy MRIs without requiring modifications to downstream pipeline architectures.

Our main contributions are summarized as follows:
\begin{itemize}
    \item We present the first comprehensive frequency-domain analysis of pathological MRI anomalies, revealing their distinctive frequency characteristics that are specific to lesion pathology.
    \item We propose FDP, a novel preprocessing framework for unsupervised anomaly detection that enhances MRI reconstruction by selectively filtering lesion-related frequencies while preserving anatomical integrity.
    \item We demonstrate that FDP consistently improves the performance of existing anomaly detection methods through comprehensive experimental validation.
    % \item We analyze the frequency-domain characteristics of MRIs and natural lesion information, highlighting the distinct patterns of low and high-frequency components associated with healthy tissue and abnormalities.
    % \item We propose a novel unsupervised anomaly detection preprocessing pipeline for MRI based on frequency decomposition, enhancing healthy MRI reconstruction and integrating seamlessly with other techniques.
    % \item We highlight the importance of high-frequency signals in MRI and integrate them into the reconstruction process, thereby enhancing structural textures and preserving anatomical details.
\end{itemize}

\begin{figure}[t]
    \centering
    \includegraphics[width=0.96\linewidth]{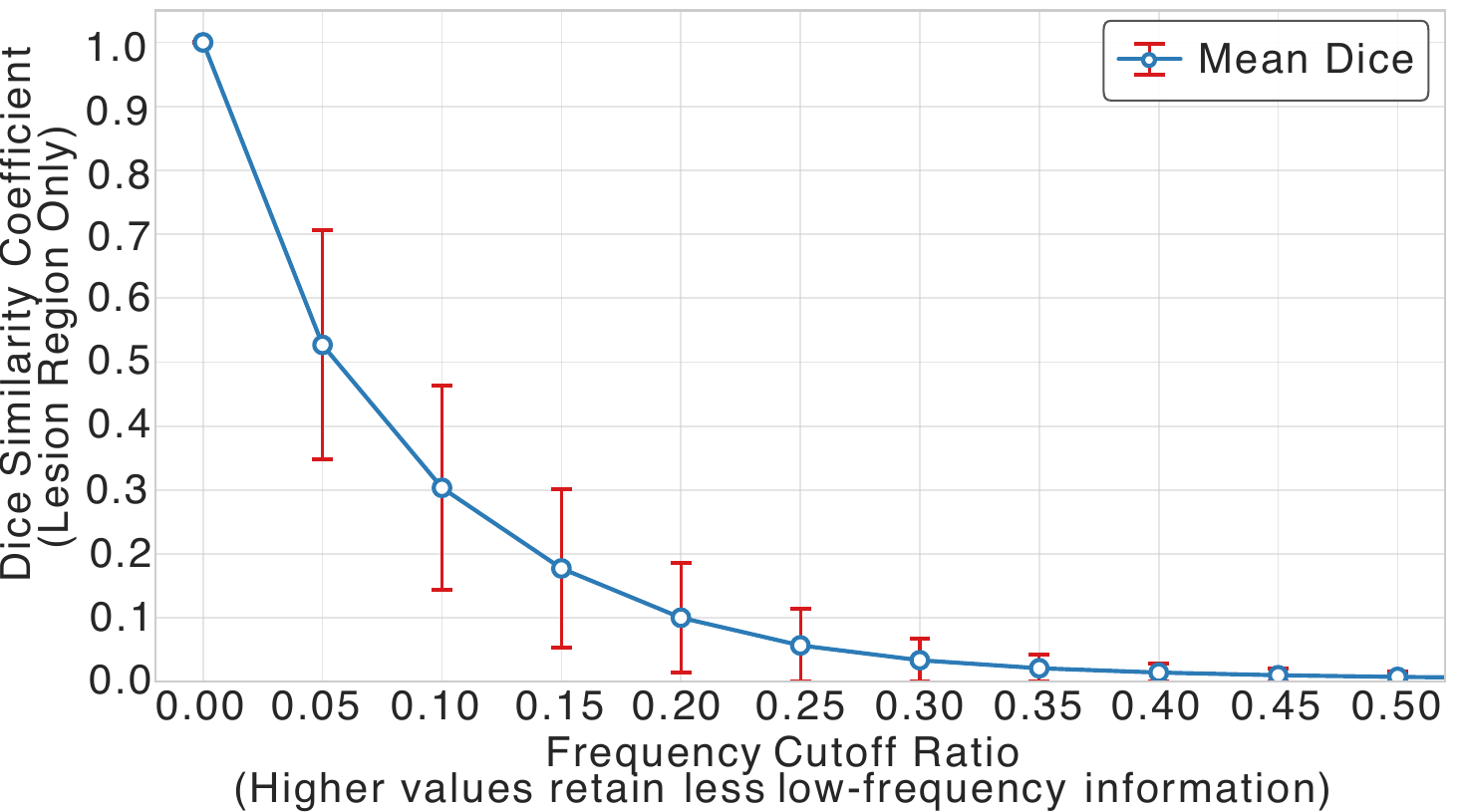}
    \caption{The effect of high-pass filter threshold on anomaly content in brain MRIs. The DICE coefficient between high-pass-filtered images and ground truth shows anomaly information primarily reside in low-frequency components.}
    \label{fig:low_freq_quant_remove}
\end{figure}

\section{Related Work}
\subsubsection{Generative Models} Recent advances in generative models have significantly impacted medical image analysis, with several architectures demonstrating particular promise. Among these, generative adversarial networks (GAN)-based~\cite{goodfellow2020generative, isola2017image} and variational autoencoder (VAE)-based~\cite{pinaya2021unsupervised, raad2023unsupervised, wijanarko2024tri} methods have gained attention for their success in image generation and translation. The emergence of Denoising Diffusion Probabilistic Models (DDPMs)~\cite{ho2020denoising, nichol2021improved} marked a significant breakthrough, achieving superior generation quality through iterative denoising processes that provide better distribution coverage and training stability compared to previous approaches. Building on these foundations, the Latent Diffusion Model (LDM)~\cite{rombach2022high} improves computational efficiency by performing the diffusion process in a compressed latent space, and has been applied to medical and heterogeneous imaging~\cite{kebaili2025multi, dong2025generative, lin2024pair}.

\subsubsection{Unsupervised Anomaly Detection in Brain MRI} Baur et al.~\cite{baur2019deep} initiated UAD in brain MRI lesion analysis by training spatially variant autoencoders on normal data only, using spatial bottlenecks to preserve spatial details, and detected lesions by analyzing pixel-wise reconstruction errors. Subsequent methods such as F-AnoGAN~\cite{schlegl2019f} and AAE~\cite{chen2018unsupervised} introduced generative adversarial networks for UAD in MRI, assuming that healthy regions remain unchanged during reconstruction and that latent representations of lesion and non-lesion images are similar. However, this assumption often fails, as lesion intensities can distort latent projections~\cite{wijanarko2024tri}.
With the rise of denoising-based techniques, denoising autoencoders (DAEs)\cite{kascenas2022denoising} and related approaches simulate anomalies by adding noise, typically in the form of discrete points. Others \cite{iqbal2023unsupervised, behrendt2024patched} apply large regular masks and reconstruct the original image from the corrupted input. AnoDDPM~\cite{wyatt2022anoddpm} introduced a partial diffusion scheme that adds noise at a specific timestep and reconstructs from the corrupted image. Recently, Tri-VAE~\cite{wijanarko2024tri} improved UAD by introducing noise while decoupling metric learning from latent sampling. It aligns images to a lesion-free distribution, incorporates a semantic-guided retrieval module, and uses structural similarity as an additional training objective. Despite their controllability, artificial noise methods often lack natural variability and anatomical structure, limiting their realism and effectiveness. 
mDDPM~\cite{iqbal2023unsupervised} applies masking in both the image and frequency domains during training to encourage a diffusion model to learn a prior over healthy anatomy. At inference, no masking is introduced, and the model is directly conditioned on the unhealthy input. As a result, the reconstruction is not explicitly constrained to suppress pathology-related cues, and the generated image may retain lesion information, which can weaken residual-based anomaly localization.

\begin{figure*}[t]
    \centering
    \includegraphics[width=\textwidth]{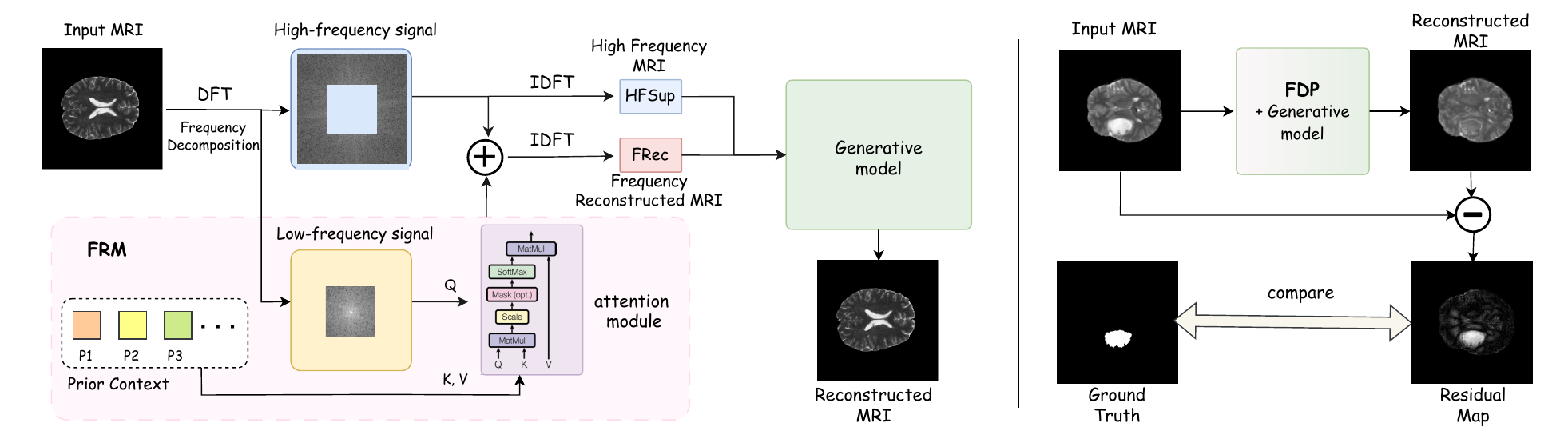}
    % \vspace{-\baselineskip}
    \caption{Training and inference pipeline of the proposed FDP method.
\textbf{Left: Training Phase.} High-frequency signals are used for both frequency-domain reconstruction and as auxiliary input (HFSup) to enhance structural details.
\textbf{Right: Inference Phase.} The input MRI with lesions is processed by FDP for frequency reconstruction, then fed into the generative model. }
    \label{fig:main_structure}
    \vspace{-\baselineskip}
    % \vspace{-6em}
\end{figure*}

\subsubsection{Fourier Transform} Fourier Transform converts a signal from the spatial domain into the frequency domain by breaking it down into its sine and cosine components at different frequencies, revealing the signal's frequency characteristics. In image processing, the Fourier Transform is used for frequency domain analysis~\cite{wang2024fremim, vaish2024fourier, dong2024shadowrefiner}. High-frequency components capture fine details, edges, and textures, while low-frequency components represent broader and regional structural features. Fourier Transform converts signals into the frequency domain, enabling the application of filters for tasks such as noise reduction, edge detection, and compression,  enhancing image processing capabilities
~\cite{cochran1967fast, ye2024diffusionedge, zhou2024muge}.
Given a 2D image $I\in R^{H\times W}$, where $I(x, y)$ denotes the image signal located at position $(x, y)$, the corresponding 2D Discrete Fourier Transform (2D-DFT) can be represented as follows:
\begin{equation}
\begin{aligned}
f(u,v) &= \mathrm{DFT}(I(x, y)) \\
&= \sum_{x=0}^{H-1}\sum_{y=0}^{W-1}I(x, y)e^{-j2\pi(\frac{ux}{H}+\frac{vy}{W})},
\end{aligned}
\end{equation}
here $u$ and $v$ serve as indices representing the horizontal and vertical spatial frequencies in the Fourier spectrum, and $f(u, v)$ denotes the corresponding frequency signal. The 2D-IDFT is defined as follows:
\begin{equation}
\begin{aligned}
I(x,y) &= \mathrm{IDFT}(f(u, v)) \\
&= \frac{1}{HW}\sum_{u=0}^{H-1}\sum_{v=0}^{W-1}f(u, v)e^{j2\pi(\frac{ux}{H}+\frac{vy}{W})}.
\end{aligned}
\end{equation}
Both DFT and IDFT can be computed using Fast Fourier Transform (FFT) algorithm \cite{brigham1967fast}.

\section{Methodology}
We start with a systematic analysis of pathological signatures in the frequency domain of brain MRI, identifying their distinct frequency signal patterns. Building on the findings, we propose a novel frequency-domain signal processing framework that selectively suppresses anomalies while preserving anatomical features.
The overall structure of Frequency-Decomposition Preprocessing (FDP) pipeline is shown in Figure~\ref{fig:main_structure}. FDP primarily consists of the Frequency Reconstruction Module (FRM) for core frequency signal reconstruction, supplemented by the High-Frequency component (HFSup) for enhanced structural detail refinement. MRIs are first transformed into the frequency domain and split into high- and low-frequency components using a frequency filter. The FRM reconstructs the low-frequency part, which is then merged with the preserved high-frequency signals to form a complete frequency-domain representation. This is converted back into the spatial domain to guide healthy MRI reconstruction using a generative model. In parallel, the high-frequency signals are also transformed into an image and used as an auxiliary structural prior, which helps preserve anatomical structures and sharpen edges.

\subsection{Characteristics of Anomaly in Frequency Domain} 
From the right part of Figure~\ref{fig:main_structure}, we observe that the lesion region in pathological MRI appears as a smooth, homogeneous area in the spatial domain. According to Fourier principles, such gradual intensity variations primarily contribute to low-frequency signal components (Additional visual comparisons are provided in the Appendix).
To verify this assumption, we transform the MRI into the frequency domain and apply a high-pass filter with threshold $m$, retaining a proportion of $(1-m)$ of the high-frequency components. Meanwhile, we employ the DICE coefficient~\cite{dice1945measures} to quantitatively assess the correlation between low-frequency components and lesion-related information. 
As shown in Figure~\ref{fig:low_freq_quant_remove}, the DICE coefficient decreases rapidly as the high-pass filtering threshold $m$ increases, dropping below 0.1 when $m$ reaches 0.2. This indicates that lesion-related signals are primarily contained in the low-frequency components and can be effectively removed through high-pass filtering.
\textbf{Unlike general anomaly detection tasks where anomalies are mainly concentrated in high-frequency components, MRI lesions typically manifest as continuous, low-frequency regional signals.} Leveraging this property, high-pass filtering not only suppresses irrelevant smooth or uniform background signals but also enhances the visibility of structural features such as edges and textures. Since high-frequency components contain minimal lesion-related information, emphasizing them allows the model to better capture anatomical structural information and improves its reconstruction fidelity.

\begin{figure*}[t]
    \centering
    \includegraphics[width=0.9\linewidth]{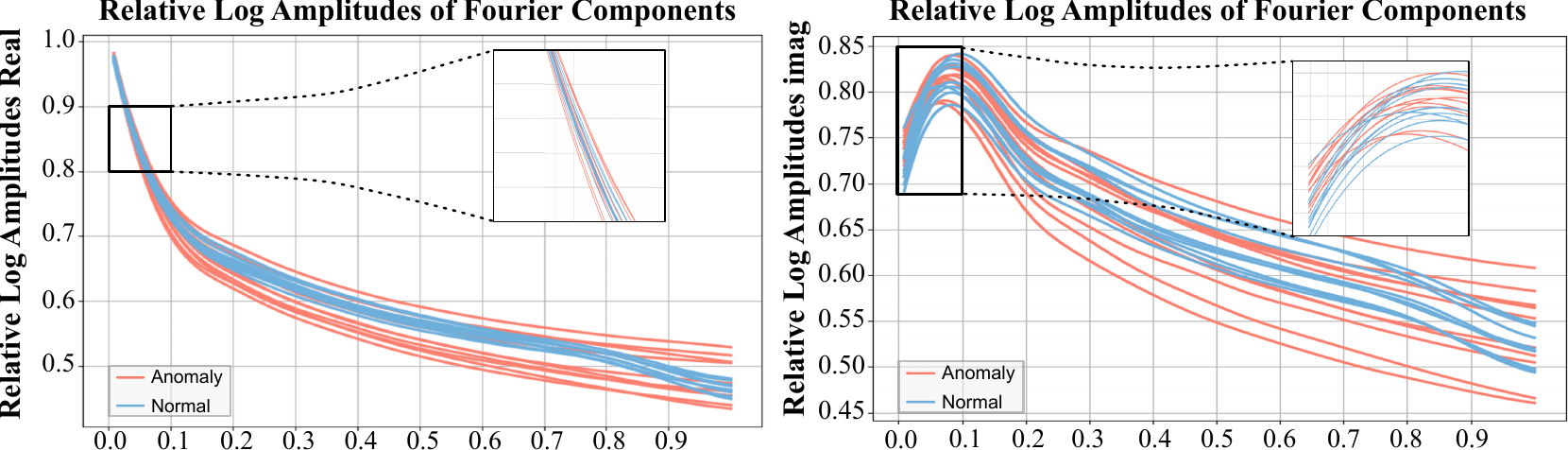}
    \caption{The horizontal and vertical axes represent the high-pass filtering threshold $m$ and the normalized real and imaginary parts of the amplitude, respectively.}
    \label{fig:intro}
\end{figure*}

\subsection{Consistency of Low Frequency Signals}
However, simply removing low-frequency signals may lead to information loss. To address this, we investigate whether it was possible to recover the lost low-frequency signals without reintroducing lesion-related information as much as possible.
Our systematic frequency-domain analysis reveals fundamental differences between normal and pathological MRIs in their real and imaginary components. As demonstrated in Figure~\ref{fig:intro}, through log-amplitude analysis in the frequency domain, two principal findings emerge: (1) normal MRIs exhibit highly consistent real-component patterns in low-frequency regions ($m \leq 0.1$), whereas pathological cases show substantially greater dispersion in the same bands; (2) both groups display comparable signal variability in high-frequency ranges ($m > 0.1$). 
These observations are consistent with basic frequency-domain principles: low-frequency components ($m \leq 0.1$) represent global anatomical structures, while high-frequency components ($m > 0.1$) capture local details. In normal MRIs, stable low-frequency patterns reflect structural homogeneity, while pathological alterations disrupt this coherence, manifesting as increased dispersion in low-frequency bands—consistent with lesion-induced anatomical discontinuities (More quantitative analysis is provided in the Appendix).

\subsection{Frequency Decomposition}
The signal processing is performed in the frequency domain. Although direct access to native k-space data is ideal for this purpose, such data are often difficult to obtain in practice. This is due to several factors, including patient privacy considerations, the large storage requirements of raw frequency data, and substantial variability in acquisition protocols across MRI vendors. Moreover, most publicly available datasets provide only reconstructed images rather than raw k-space measurements. Therefore, we apply a Fourier transform to convert MRI images back into the frequency domain for subsequent processing.

For a MRI $I\in R^{H\times W}$, we apply 2D Discrete Fourier Transform (2D-DFT) to convert the image from the spatial domain to the frequency domain. We then use an ideal high-pass filter $\mathrm{H_{hp}}(u, v)$ to retain the high-frequency components and remove the low-frequency components. The filter $\mathrm{H_{hp}}(u, v)$ is defined as follows:
\begin{equation}
\begin{aligned}
\mathrm{H_{hp}}(u, v) =
\begin{cases}
0 & \text{if } \mathrm{D}(u, v) \leq \mathfrak{D}_0 \\
1 & \text{if } \mathrm{D}(u, v) > \mathfrak{D}_0
\end{cases},
\end{aligned}
\end{equation}
where \( \mathrm{D}(u, v) \) is the distance from the frequency origin and  \( \mathfrak{D}_0 \) refers to the distance from the current point to the center of the frequency domain, given by:
\begin{align}
\mathrm{D}(u, v) &= \sqrt{(u - H/2)^2 + (v - W/2)^2}, 
\end{align}
\begin{align}
\mathfrak{D}_0 &= \min(m*H, m*W), m \in [0,1], 
\end{align}
where $m$ represents the threshold of the high-pass filter.

Finally, the 2D Inverse Discrete Fourier Transform (2D-IDFT) converts the filtered signal back from the frequency domain to the spatial domain, i.e.,
\begin{align}
f(u, v) & = \mathrm{DFT}(I(x, y)),
\end{align}
\begin{align}
f_\mathrm{h}(u, v) &= f(u, v) \odot \mathrm{H_{hp}}(u, v), 
\end{align}
\begin{align}
f_\mathrm{l}(u, v) &= \mathrm{CROP}(f(u, v), f_\mathrm{h}(u, v)), 
\end{align}
\begin{align}
I_\mathrm{h} &= \mathrm{IDFT}(f_\mathrm{h}(u, v)),
\end{align}
here $\mathrm{CROP}(a, b)$ means to delete off the high-frequency signals $b$ from the complete frequency signals $a$. 
In this way, we decompose an MRI $I$ into low-frequency signals $f_\mathrm{l}$ and high-frequency signals $f_\mathrm{h}$ using a high-pass filter. In addition, the high-frequency image $I_{h}$ can later be used to enhance image detail quality.

\subsection{Frequency Reconstruction Module}
Through comprehensive quantitative analyses (e.g., PCA, t-SNE, and maximum likelihood estimation) presented in the Appendix, we demonstrate that these low-frequency signals exhibit low variance and lie approximately on a low-dimensional manifold. This motivates us to model the distribution of low-frequency signals using a prior context bank $\boldsymbol{P}$, estimated from healthy MRIs, such that the low-frequency component of any healthy MRI can be approximated via latent sampling and mapping. 

We therefore propose the Frequency Reconstruction Module (FRM), which serves as a decoder or a linear mapping based on learned dictionaries. This is analogous to the reparameterization trick used in variational inference, enabling us to sample smooth, consistent low-frequency signals in a differentiable manner.
We adopt an attention-based retrieval strategy from a learnable set of prior context $\boldsymbol{P} = [p_{1}, p_{2}, \dots p_{k}]$, where $k$ denotes the number of learnable prior context elements and each $p_{i} \in R^{(m*H)\times (m*W)}$. These contexts are initialized using k-means++~\cite{arthur2006k} clustering over the training set to promote faster and more stable convergence. Given a low-frequency query $f_\mathrm{l}$, we reconstruct the MRI via:
\begin{align}
\hat{f_\mathrm{l}} &= \mathrm{ATTN}(f_\mathrm{l}, P, P),
\end{align}
\begin{align}
\hat{f} &= \mathrm{MERGE}(\hat{f_\mathrm{l}}, f_\mathrm{h}),
\end{align}
\begin{align}
\hat{I} &= \mathrm{IDFT}(\hat{f}),
\end{align}
where $\mathrm{ATTN}$ represents an attention module, and $\mathrm{MERGE}(a,b)$ merges the low-frequency component $a$ and the high-frequency component $b$ into a complete frequency representation. In this way, the reconstructed low-frequency signals $\hat{f}\mathrm{l}$ are integrated with the original high-frequency signals $f\mathrm{h}$ to obtain the reconstructed frequency signals $\hat{f}$, which are subsequently transformed back to the spatial domain to produce the frequency-reconstructed MRI $\hat{I}$. We supervise the reconstruction of the low-frequency signals using the $\mathrm{L1}$ loss, formulated as $\mathrm{L1}(\hat{f}\mathrm{l}, f\mathrm{l})$.
This design enables the model to generate plausible low-frequency content conditioned on the preserved high-frequency signal and the prior context, thereby minimizing the chance that lesion-related information is reintroduced during reconstruction.

Subsequently, $\hat{I}$ and $I_\mathrm{h}$ can be used for MRI reconstruction using generative models such as VAEs and LDMs. Additionally, the obtained $I_\mathrm{h}$ can serve as auxiliary structural information, which we term High-Frequency Supplement (HFSup), to enhance anatomical detail preservation.

\section{Experiments}
\subsection{DataSets}
We adopted T2-weighted MRI as the primary modality because it is widely available and provides strong tissue contrast. In comparison, T1-weighted images primarily capture anatomical detail but often show weaker conspicuity for many lesions. FLAIR acquisition protocols are less standardized across sites, and its advantage can be region dependent.

For training, we used the publicly available IXI dataset, which includes 560 T1- and T2-weighted brain MRI scans from three clinical centers. The dataset comprises only healthy subjects, providing a clean distribution for modeling normal anatomy.
For evaluation, we conduct 10 evaluation runs, each using 32 cases randomly sampled from BraTS 2020 dataset~\cite{bakas2018identifying, menze2014multimodal}, which includes 369 annotated brain MRI scans across four modalities (T1, T1-CE, T2, FLAIR) and corresponding tumor segmentations. Each scan contains roughly 155 slices, with T2 images acquired at $0.9375 \times 0.9375~\text{mm}^{2}$ in-plane resolution, $0.125~\text{mm}$ slice thickness, and $240 \times 240$ image size.

In addition, to further validate the generalization capability of our model,  we conducted experiments on the T2-weighted modality across several additional MRI brain datasets, including the Multimodal Brain Tumor Segmentation Challenge 2021 (BraTS21)~\cite{baid2021rsna}, the multiple sclerosis dataset (MSLUB)~\cite{lesjak2018novel}, and the Multiple Sclerosis Lesion (MSSEG-2)~\cite{commowick2021msseg}. These datasets use similar storage formats as BraTS20 but cover more diverse brain pathologies, offering broader scenarios to evaluate model generalization.

\subsection{Implementation Details}
We applied skull stripping with HD-BET~\cite{isensee2019automated} to filter out the regions belonging to the foreground area so that the masking block can only be applied to the foreground pixel patches. 
During the experiments, all slices are resized to a uniform resolution of $256 \times 256$ after normalization.
Unless otherwise specified, the LDM was chosen as the default generative model, the number of prior context was set to 128 by default, and the high-pass filtering threshold $m$ used for FRM and HFSup are both set to $0.10$ as default.
The model was trained on 4 NVIDIA V100 GPUs (32GB) using the Adam optimizer, with a learning rate of 2e-5 and a batch size of 32 for 800 epochs.
The details of evaluation metrics (DICE, AUPRC, AUROC) and post-processing can be found in Appendix. All of our experiments adopted the unified post-processing method.

\renewcommand{\arraystretch}{1.2}
\begin{table*}[t]
\small
\centering
\resizebox{0.95\linewidth}{!}{
\begin{tabular}{llll}
\toprule
\textbf{Model}  & \textbf{DICE}   & \textbf{AUPRC}   & \textbf{AUROC}                \\ \hline
\multicolumn{4}{l}{\textbf{Base Models}} \\
VAE~\cite{chen2020unsupervised} & $34.90 \pm 2.1$ & $29.95 \pm 3.3$ & $94.46 \pm 0.5$ \\
LDM~\cite{rombach2022high} & $35.02 \pm 1.8$ & $30.75 \pm 2.7$ & $91.62 \pm 1.1$ \\
AnoDDPM~\cite{wyatt2022anoddpm} & $36.19 \pm 1.5$ & $32.01 \pm 2.4$ & $91.37 \pm 1.3$ \\
F-AnoGAN~\cite{schlegl2019f} & $37.68 \pm 1.2$ & $35.05 \pm 1.9$ & $91.88 \pm 0.9$ \\
% ASC-Net~\cite{dey2021asc} & $38.18 \pm 0.8$ & $36.44 \pm 1.5$ & $96.50 \pm 0.3$ \\
% mDDPM~\cite{iqbal2023unsupervised} & $37.80 \pm 1.1$ & $34.40 \pm 2.1$ & $92.25 \pm 0.7$ \\ % \hline
DAE (simplex)~\cite{kascenas2022denoising} & $37.38 \pm 1.4$ & $32.93 \pm 2.8$ & $94.65 \pm 0.6$ \\
% Tri-VAE (simplex)~\cite{wijanarko2024tri} & $40.47 \pm 1.0$ & $39.65 \pm 1.7$ & $95.90 \pm 0.4$ \\
DAE (coarse) & $56.87 \pm 3.2$ & $43.23 \pm 2.5$ & $95.71 \pm 0.5$ \\
% Tri-VAE (coarse) & $60.58 \pm 2.8$ & $46.15 \pm 2.2$ & $96.82 \pm 0.3$ \\ 
pDDPM~\cite{behrendt2024patched} & $46.15 \pm 2.4$ & $45.67 \pm 2.9$ & $92.01 \pm 0.9$ \\
\hline
\multicolumn{4}{l}{\textbf{FDP-enhanced Models}} \\
FDP + VAE & 
$46.32 \pm 1.9$ \textcolor{red}{(\underline{+11.42})} & 
$41.32 \pm 2.6$ \textcolor{red}{(\underline{+11.37})} & 
$92.16 \pm 0.8$ \textcolor{blue}{(\underline{-2.30})} \\

FDP + LDM & 
$52.66 \pm 2.4$ \textcolor{red}{(\underline{+17.63})} & 
$51.67 \pm 3.1$ \textcolor{red}{(\underline{+20.92})} & 
$93.12 \pm 0.9$ \textcolor{red}{(\underline{+1.50})} \\

FDP + AnoDDPM & 
$48.24 \pm 1.7$ \textcolor{red}{(\underline{+12.05})} & 
$47.56 \pm 2.9$ \textcolor{red}{(\underline{+15.55})} & 
$92.97 \pm 1.0$ \textcolor{red}{(\underline{+1.60})} \\

FDP + F-AnoGAN & 
$52.96 \pm 2.1$ \textcolor{red}{(\underline{+15.28})} & 
$50.70 \pm 2.7$ \textcolor{red}{(\underline{+15.65})} & 
$93.84 \pm 0.7$ \textcolor{red}{(\underline{+1.96})} \\

FDP + DAE (simplex) & 
$54.77 \pm 2.3$ \textcolor{red}{(\underline{+17.39})} & 
$51.88 \pm 3.0$ \textcolor{red}{(\underline{+18.95})} & 
$94.69 \pm 0.6$ \textcolor{red}{(\underline{+0.04})} \\

FDP + DAE (coarse) & 
$63.03 \pm 3.5$ \textcolor{red}{(\underline{+6.16})} & 
$61.41 \pm 3.3$ \textcolor{red}{(\underline{+18.18})} & 
$93.95 \pm 1.2$ \textcolor{blue}{(\underline{-1.76})} \\

FDP + pDDPM & 
$54.09 \pm 1.9$ \textcolor{red}{(\underline{+8.04})} & 
$51.03 \pm 3.0$ \textcolor{red}{(\underline{+5.36})} & 
$93.72 \pm 1.0$ \textcolor{red}{(\underline{+1.62})} \\

\bottomrule
\end{tabular}
}
\vspace{1ex}
\caption{Comparing anomaly detection performance on the BraTS20 dataset. Values in parentheses indicate performance changes after integrating FDP. \textcolor{red}{Red} indicates improved performance, while \textcolor{blue}{blue} indicates reduced performance.}
\label{tab:main_compare}
\end{table*}
\renewcommand{\arraystretch}{1}

\subsection{Results and Analysis}
\begin{figure*}[th]
    \centering
    \includegraphics[width=0.7\linewidth]{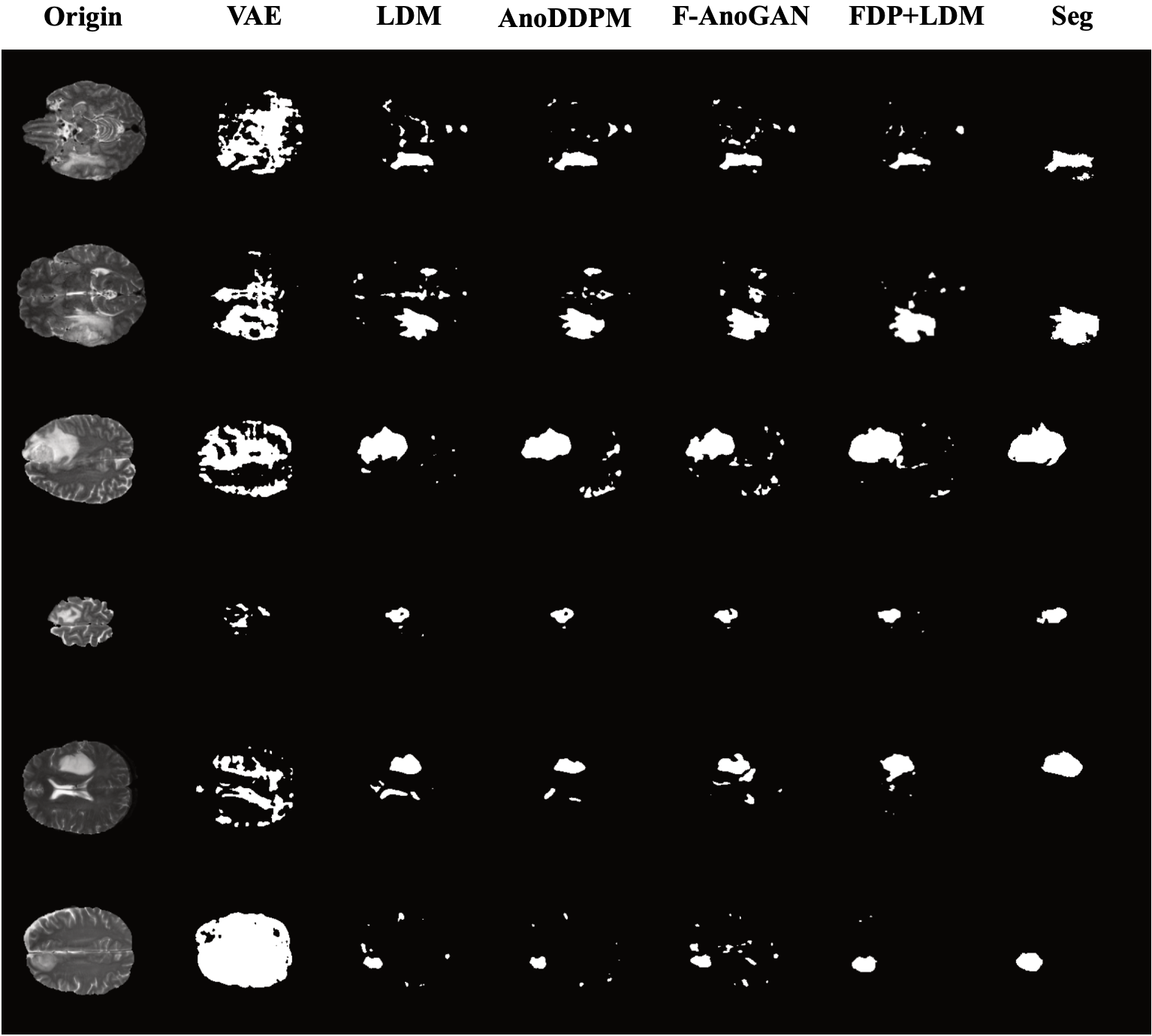}
    \caption{Visual comparison of results with other methods. Origin means the input MRIs, and Seg denotes the ground truth.}
    \label{fig:result_compare}
    \vspace{-2mm}
\end{figure*}

We evaluated our method's compatibility with other open-source approaches.
As presented in Table~\ref{tab:main_compare}, integrating FDP with the base LDM results in a 17.63\% improvement in the DICE score and a 20.92\% improvement in the AUPRC score.
Furthermore, when FDP is integrated with other methods, the results improve by at least 6.16\% in DICE and 11.37\% in AUPRC.
In terms of AUROC, our improvements are limited, with some methods exhibiting slight performance drops after integration. We attribute this to the task's complexity and the trade-offs in anomaly removal. 
As shown in Figure~\ref{fig:result_compare}, our predictions contain fewer errors, while the predicted lesion regions tend to be slightly smaller than those of other methods and the ground truth, indicating that our method achieves higher precision at the cost of slightly lower recall.
To intuitively demonstrate FDP’s effectiveness in removing MRI anomalies, we include visualizations of results at each pipeline stage in the Appendix.

\begin{table}[t]
  \centering
  \small
  \resizebox{0.85\linewidth}{!}{
  \begin{tabular}{llrrr}
    \toprule
    Dataset     & Model       & DICE    & AUPRC    & AUROC \\ 
    \midrule
    \multirow{2}{*}{BraTS21}  
                & LDM         & 29.53   & 27.52    & 92.26    \\
                & LDM+FDP     & 45.06   & 37.17    & 93.13  \\
    \midrule
    \multirow{2}{*}{MSLUB}  
                & LDM         & 8.35    & 8.62     & 84.98\\
                & LDM+FDP     & 13.06   & 14.28    & 87.14 \\
    \midrule
    \multirow{2}{*}{MSSEG-2} 
                & LDM         & 20.15   & 25.67    & 85.22  \\
                & LDM+FDP     & 34.63   & 37.29    & 88.32  \\
    \bottomrule
  \end{tabular}
  }
  \caption{Comparison of models across different datasets.}
  \label{tab:dataset_compare}
\end{table}

\begin{table}[t]
  \centering
  \small
  \resizebox{0.8\linewidth}{!}{
  \begin{tabular}{ccccc}
    \toprule
    FRM & HFSup & DICE & AUPRC & AUROC \\
    \midrule
         &       & 35.02 & 30.75 & 91.62 \\
         & \checkmark & 42.18 & 40.55 & 92.25 \\
    \checkmark &       & 50.00 & 45.86 & 92.97 \\
    \checkmark & \checkmark & 52.66 & 51.67 & 93.12 \\
    \bottomrule
  \end{tabular}
  }
  \caption{Ablation results on the impact of FRM and HFSup.}
  \label{tab:ablation}
\end{table}

Furthermore, our method exhibits strong generalization across multiple MRI datasets, achieving substantial improvements over the baseline LDM. As shown in Table~\ref{tab:dataset_compare}, it yields significant performance gains of 15.53\% and 14.48\% DICE points on standard benchmarks (BraTS21 and MSSEG-2, respectively). Notably, even for the more challenging MSLUB dataset, it maintains a robust improvement of 4.71\% DICE points. These results collectively demonstrate the consistent effectiveness of our method across diverse MRI data domains.

\subsection{Ablation Studies}
In this section, we analyze the impact of $m_\mathrm{FRM}$ ($m$ used for FRM), $m_\mathrm{HFSup}$ ($m$ used for HFSup), and the number of prior context on model performance. We further investigate how different hyperparameter settings for each component affect the overall performance. By default, we employ LDM without noise-adding as our generative model, with $m_\mathrm{FRM}$ set to 0.10, $m_\mathrm{HFSup}$ also set to 0.10, and the number of prior context set to 128.

As shown in Table~\ref{tab:dataset_compare} and Table~\ref{tab:ablation}, high-frequency information alone does not substantially improve performance. Instead, it serves as a supplementary feature that enhances the baseline model.
Relying exclusively on high-frequency information may lead to the loss of essential details in the input image, providing insufficient information for accurate MRI reconstruction.
The FRM plays a pivotal role in enhancing overall model performance by effectively suppressing lesion-related information while preserving structural details in high-frequency components, thereby improving image reconstruction quality. The synergistic integration of these modules leads to significant performance gains.

\subsubsection{High Frequency Filtering Threshold}
To determine the appropriate $m_\mathrm{FRM}$ and $m_\mathrm{HFSup}$ values for FRM and HFSup, we evaluated different mask ratios on each to assess their respective optimal settings. 
The results are presented in Table~\ref{tab:mfrm} and Table~\ref{tab:mhfsup}, respectively. In Table~\ref{tab:mfrm}, $m_\mathrm{FRM}$ is the variable parameter, while $m_\mathrm{HFSup}$ is fixed at $0.10$. Conversely, in Table~\ref{tab:mhfsup}, $m_\mathrm{HFSup}$ is the variable parameter, and $m_\mathrm{FRM}$ is fixed at $0.10$.

\begin{table}[t]
\centering
\begin{subtable}[b]{0.49\linewidth}
\centering
\resizebox{\linewidth}{!}{
\begin{tabular}{cccc}
\toprule
$m_\mathrm{FRM}$  & DICE & AUPRC & AUROC \\
\midrule
0.01 & 39.45 & 37.13 & 90.54  \\
0.05 & 50.00 & 44.86 & 92.97  \\
0.10 & 52.66 & 51.67 & 93.12 \\
0.15 & 45.24 & 43.56 & 91.78  \\
0.20 & 43.89 & 41.87 & 93.08  \\
0.25 & 43.75 & 41.68 & 92.73 \\
0.30 & 41.82 & 39.02 & 92.45 \\
\bottomrule
\end{tabular}
}
\caption{Effects of $m_\mathrm{FRM}$ (with $m_\mathrm{HFSup}=0.10$).}
\label{tab:mfrm}
\end{subtable}
\hfill
\begin{subtable}[b]{0.49\linewidth}
\centering
\resizebox{\linewidth}{!}{
\begin{tabular}{cccc}
\toprule
$m_\mathrm{HFSup}$ & DICE & AUPRC & AUROC \\
\midrule
0.01 & 38.27 & 38.48 & 90.01  \\
0.05 & 50.13 & 49.04 & 92.88  \\
0.10 & 52.66 & 51.67 & 93.12 \\
0.15 & 51.31 & 42.78 & 93.82  \\
0.20 & 49.44 & 49.47 & 92.67  \\
0.25 & 49.93 & 48.64 & 93.85 \\
0.30 & 50.76 & 49.49 & 93.75 \\
\bottomrule
\end{tabular}
}
\caption{Effects of $m_\mathrm{HFSup}$ (with $m_\mathrm{FRM}=0.10$).}
\label{tab:mhfsup}
\end{subtable}
\caption{Comparison of hyperparameter sensitivity for FRM.}
\label{tab:hyperparam-comparison}
\end{table}

As shown in Table~\ref{tab:mfrm}, when $m_\mathrm{FRM}$ is set to 0.10, the model achieves optimal performance by effectively removing lesions while retaining high-frequency MRI information and supplementing low-frequency signals through reconstruction.
When $m_\mathrm{FRM}$ is set to 0.01, the low-frequency signals primarily capture general image characteristics (e.g., illumination intensity, etc.), which are common across similar images. In this case, the method fails to remove lesions effectively during inference, rendering it ineffective.
When $m_\mathrm{FRM}$ is too large, the frequency representation mainly encodes subject-specific features, causing the signals to become overly dispersed and difficult to simulate using the prior context. This leads to substantial deviations between the generated and original images, resulting in suboptimal performance.

As shown in the Table~\ref{tab:mhfsup}, when $m_\mathrm{HFSup}$ is set to 0.01, lesion information is directly introduced due to the overly low threshold. The retention of excessive information shifts the model's focus toward high-frequency signals containing lesion details, leading to degraded performance. Conversely, when $m_\mathrm{HFSup}$ exceeds 0.15, the high-frequency structural information becomes insufficient, markedly reducing its contribution to model performance.

\subsubsection{Prior Context Number}

\begin{table}[!b]
\centering
\resizebox{0.8\linewidth}{!}{
\begin{tabular}{cccc}
\toprule
Prior Context Number & DICE & AUPRC & AUROC \\
\midrule
16  & 45.94 & 45.23 & 92.56  \\
32  & 47.55 & 47.29 & 94.13 \\
64  & 50.31 & 49.70 & 92.78  \\
128 & 52.66 & 51.67 & 93.12  \\
256 & 52.74 & 50.56 & 93.01  \\
\bottomrule
\end{tabular}
}
\caption{Effects of different numbers of prior context.}
\label{tab:ablation_prior_context_number}
\end{table}

We conducted experiments with varying amounts of prior context, evaluating five configurations (16, 32, 64, 128, and 256) to identify the optimal configuration for reconstructing low-frequency signals of healthy MRI images. As shown in Table~\ref{tab:ablation_prior_context_number}, increasing the amount of prior context improves performance, with gains plateauing around 128. The initial addition of 16 contexts yields about a 2\% improvement, but further increases bring limited benefit due to empty boundary slices and high inter-slice similarity. Given that typical MRI scans contain approximately 155 slices, 128 contexts effectively capture the volume's low-frequency structure without incurring unnecessary computational overhead.

\section{Conclusion}
In this paper, we present the first systematic frequency-domain analysis of pathological signatures in brain MRI, revealing two key properties: (1) the frequency separability between lesions and normal anatomy, and (2) the consistent low-frequency representations in healthy brain MRIs. Motivated by these insights, we propose Frequency-Decomposition Preprocessing (FDP), the preprocessing framework for UAD that leverages frequency-domain reconstruction to suppress pathologies and preserve anatomical structures. FDP integrates seamlessly with existing methods, without requiring any architectural changes to the original model. Through extensive experiments, we show that frequency-domain processing effectively suppresses pathological information, providing a reliable solution for unsupervised anomaly detection in brain MRI.
We hope that our method will inspire new research directions for UAD in MRI and contribute to advancing the field.

\section{Acknowledgments}
This work was supported by National Natural Science Foundation of China (Grant No. 62371409) and Fujian Provincial Natural Science Foundation of China (Grant No. 2023J01005).

\bibliography{aaai2026}

%%%%%%%%%%%%%%%%%%%%%%%%%%%%%%%%%%%%%%%%%%%%%%%%%%%%%%%%%%%%

\newpage

\appendix

\section{FDP: A Frequency-Decomposition Preprocessing Pipeline for
Unsupervised Anomaly Detection in Brain MRI \\ 
Technical Appendices and Supplementary Material}

\begin{figure*}[!t]
  \centering
  \begin{subfigure}[b]{0.32\textwidth}
    \includegraphics[width=\linewidth]{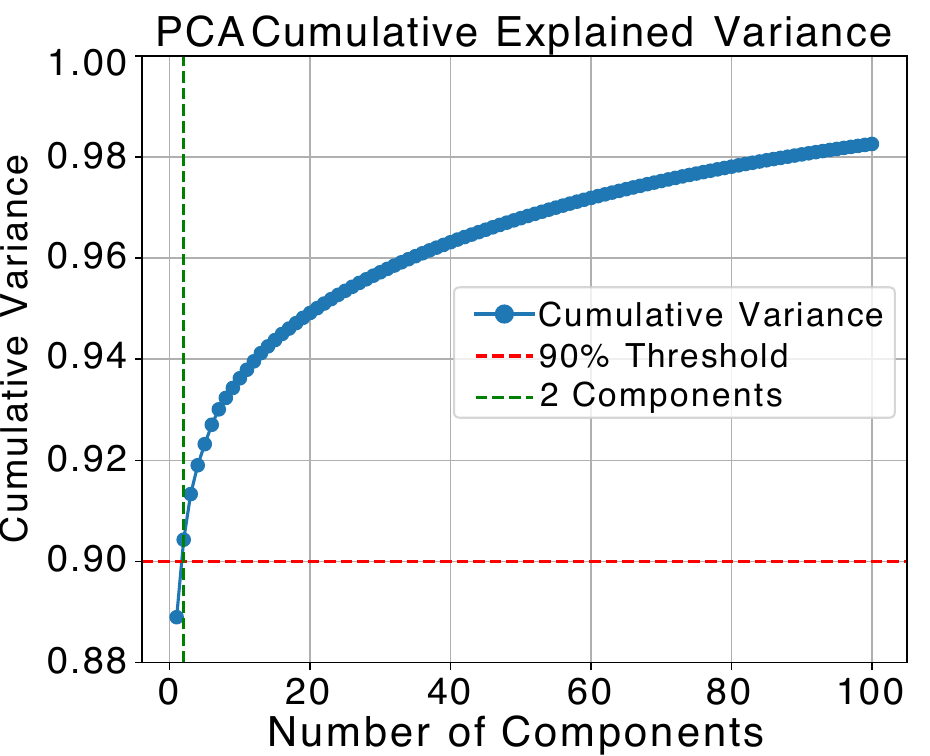}
    \caption{Variance Distribution via PCA.}
    \label{fig:pca}
  \end{subfigure}
  \hfill
  \begin{subfigure}[b]{0.32\textwidth}
    \includegraphics[width=\linewidth]{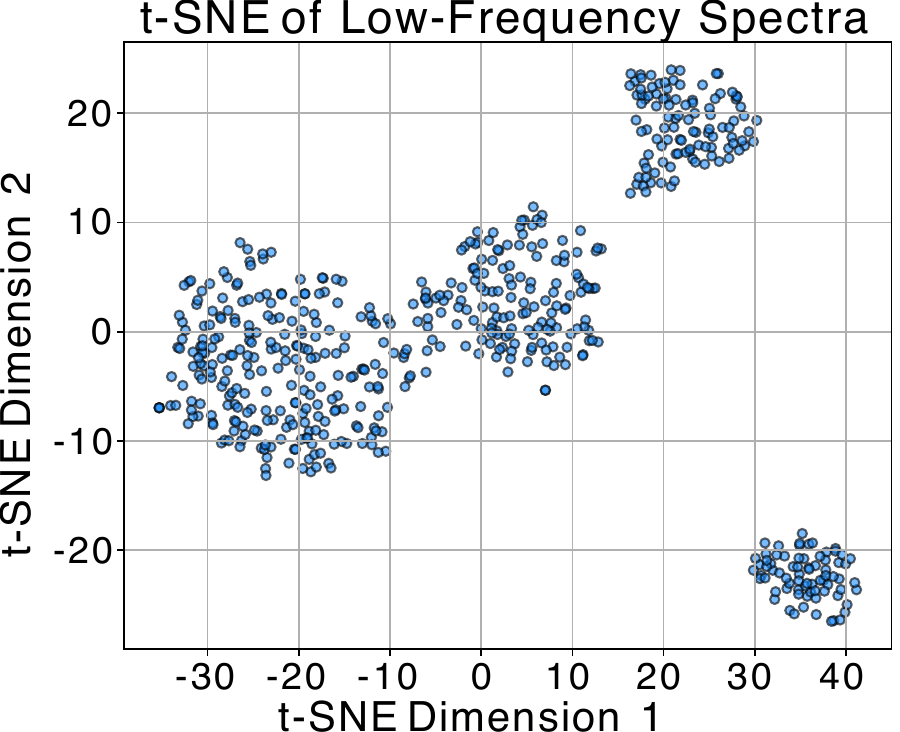}
    \caption{Clustering and Visualization via t-SNE.}
    \label{fig:tsne}
  \end{subfigure}
  \hfill
  \begin{subfigure}[b]{0.32\textwidth}
    \includegraphics[width=\linewidth]{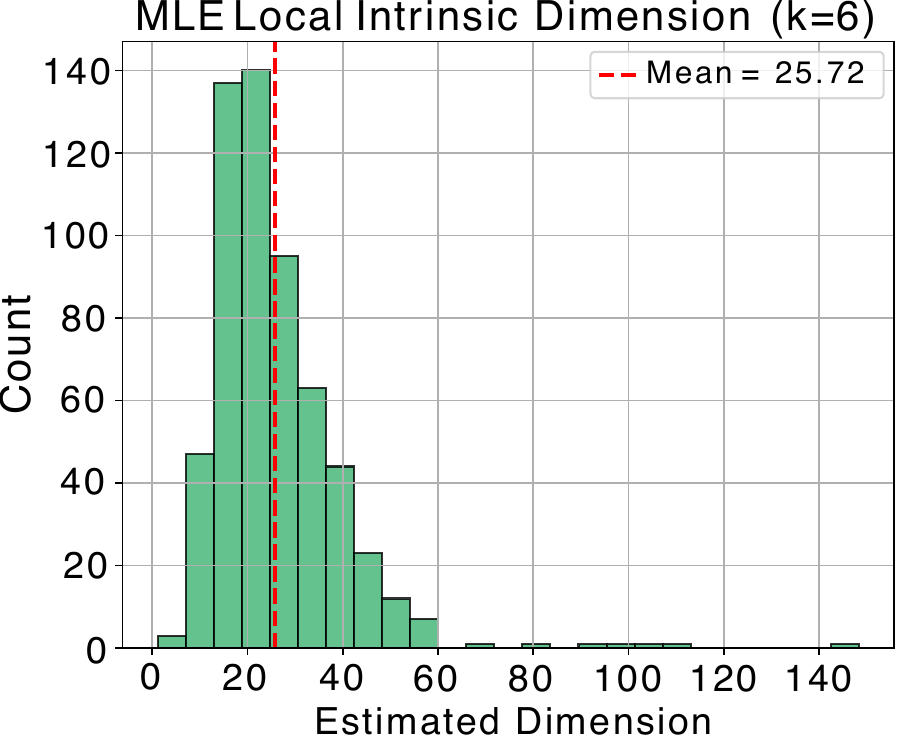}
    \caption{Intrinsic Dimensionality Analysis.}
    \label{fig:mle}
  \end{subfigure}
  \caption{Low-Frequency Signals Quantitative Analysis.}
  \label{fig:low_freq_quant}
\end{figure*}

\subsection{Low-Frequency Signal Quantitative Analysis}
\label{quantitative_analysis}
To validate the assumption that healthy brain MRIs share a consistent low-frequency information, we conduct a multi-faceted statistical analysis, as shown in Figure~\ref{fig:low_freq_quant}. Specifically, we applied a relatively lenient low-pass filter with a threshold of $m=0.2$, and performed quantization analysis on the low-frequency signal component filtered by this threshold.

We applied Principal Component Analysis (PCA)~\cite{abdi2010principal} to the low-frequency information extracted from healthy MRI slices. As shown in Figure~\ref{fig:pca}, As shown in Figure~\ref{fig:pca}, the first two principal components together account for over 90\% of the total variance, indicating a highly concentrated global structure. This suggests that most of the variability in the low-frequency domain can be effectively captured within a compact latent space.

Additionally, we projected the low-frequency information into a 2D space using t-SNE~\cite{maaten2008visualizing}. As shown in Figure~\ref{fig:tsne}, the visualization reveals four clear clusters in the embedding space, indicating that, despite the overall global consistency of the data, there exist nontrivial variations across subgroups. These clusters may be associated with demographic or scanner-related factors, suggesting that the low-frequency domain is organized into multiple coherent substructures.

Furthermore, we estimated the intrinsic dimensionality of the low-frequency space using the Maximum Likelihood Estimation (MLE)–based method~\cite{levina2004maximum}. As shown in Figure~\ref{fig:mle}, the estimated local dimensions are mostly concentrated around a mean of approximately 25.72, with a noticeable spread across samples. This distribution indicates a moderate level of local complexity and structural diversity, consistent with the hypothesis that the low-frequency space does not form a single rigid manifold but instead comprises smoothly varying submanifolds.

Together, these findings justify the design of our learnable prior module (FRM), which captures both global coherence and local diversity in healthy low-frequency features.

\subsection{Frequency-Domain Visualization Analysis of Anomalies}
\label{visual_motivate}
\begin{figure*}[htbp]
% \vspace{1ex}
    \centering
    \noindent
    \hbox to 0.96\textwidth{%
        \makebox[0.14\textwidth][c]{\textbf{Origin}}%
        \makebox[0.14\textwidth][c]{\textbf{0.01}}%
        \makebox[0.14\textwidth][c]{\textbf{0.05}}%
        \makebox[0.14\textwidth][c]{\textbf{0.10}}%
        \makebox[0.14\textwidth][c]{\textbf{0.20}}%
        \makebox[0.14\textwidth][c]{\textbf{0.30}}%
        \makebox[0.14\textwidth][c]{\textbf{Seg}}%
    }
    \vbox{

    \centering
    \includegraphics[width=0.96\textwidth]{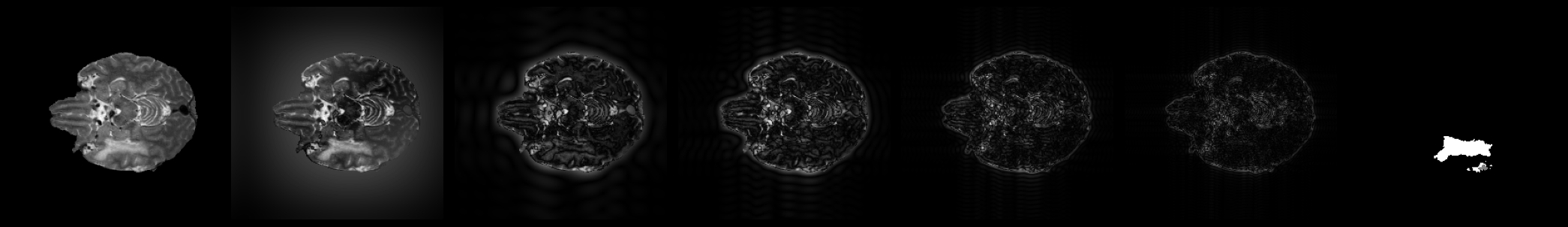}
    \vspace{-\baselineskip}
    % \vspace{-1ex}
    
    \centering
    \includegraphics[width=0.96\textwidth]{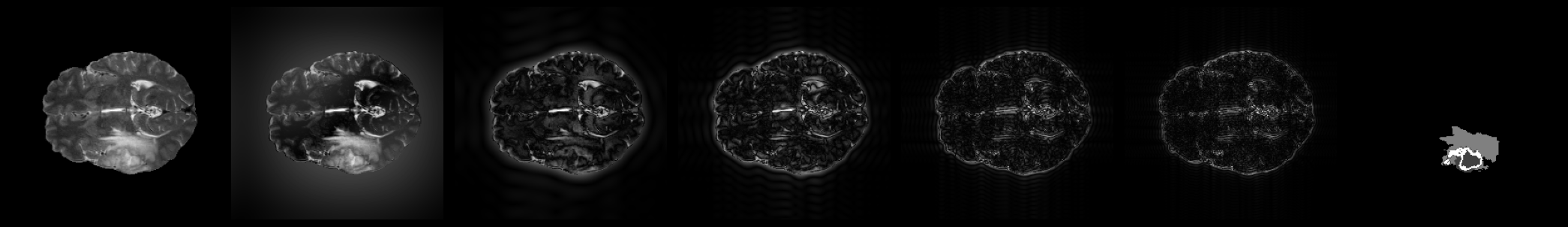}
    \vspace{-\baselineskip}
    % \vspace{-1ex}
    
    \centering
    \includegraphics[width=0.96\textwidth]{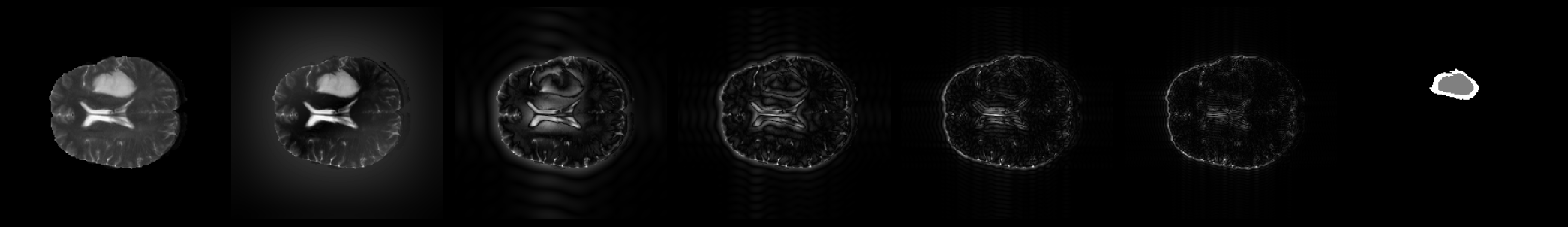}
    \vspace{-\baselineskip}
    % \vspace{-1ex}
    
    \centering
    \includegraphics[width=0.96\textwidth]{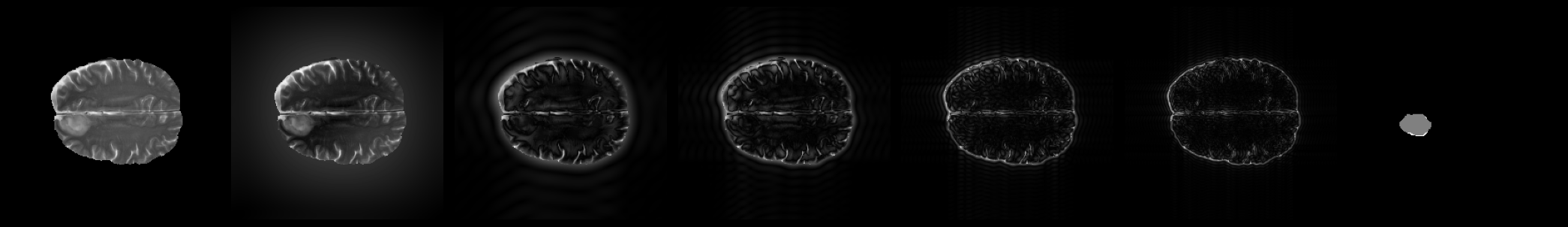}
    \vspace{-\baselineskip}
    % \vspace{-1ex}
    
    \centering
    \includegraphics[width=0.96\textwidth]{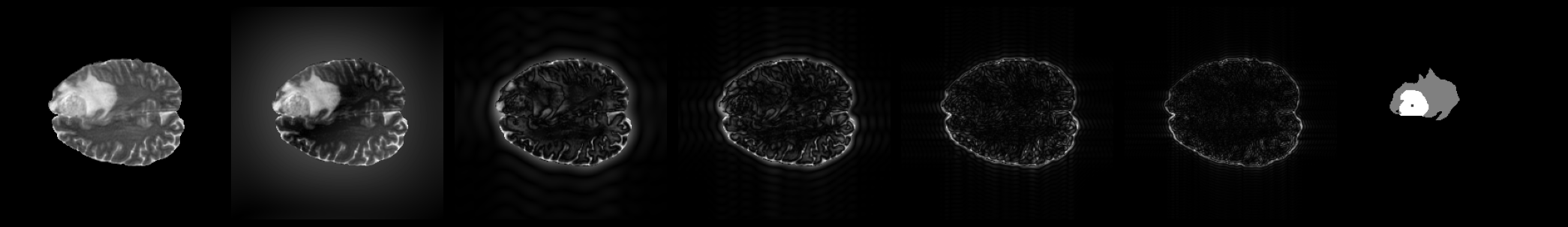}
    \vspace{-\baselineskip}
    % \vspace{-1ex}
    
    \centering
    \includegraphics[width=0.96\textwidth]{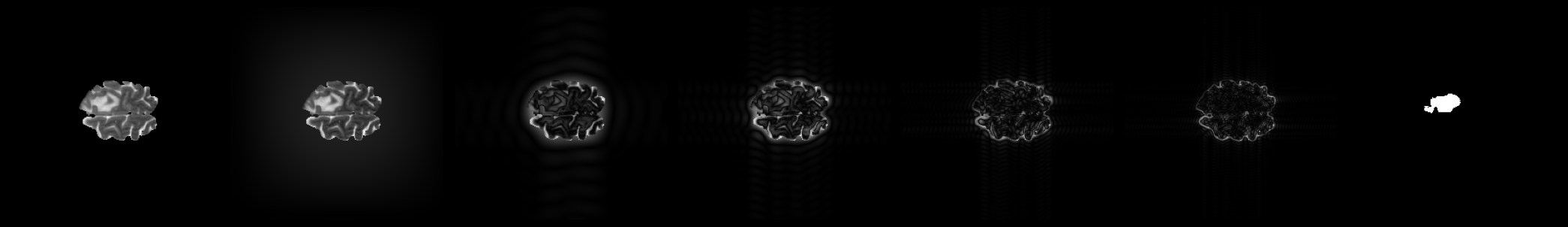}
    \vspace{-\baselineskip}
    }
    \vspace{1ex}
    \caption{The original MRIs containing anomalies and the corresponding MRIs after high-pass filtering with thresholds 0.01, 0.05, 0.10, 0.20, and 0.30 are displayed. The high-pass filtering thresholds represent the proportion of the overall frequency-domain signal area masked by the filter. The last column shows the ground-truth anomaly regions.}
    \label{fig:mask_ratio_show}
\end{figure*}

As shown in Figure~\ref{fig:mask_ratio_show}, lesion information in MRIs, characterized by smooth, continuous spatial profiles, is progressively attenuated as the high-pass filtering threshold increases, whereas structural details such as anatomical boundaries are largely preserved up to a certain threshold.

\begin{figure*}[htbp]
% \vspace{1ex}
    \centering
    \noindent
    \hbox to 0.96\textwidth{%
        \makebox[0.12\textwidth][c]{\textbf{Origin}}%
        \makebox[0.12\textwidth][c]{\textbf{RemoveLF}}%
        \makebox[0.12\textwidth][c]{\textbf{LFRec}}%
        \makebox[0.12\textwidth][c]{\textbf{FRec}}%
        \makebox[0.12\textwidth][c]{\textbf{Recon}}%
        \makebox[0.12\textwidth][c]{\textbf{Residual}}%
        \makebox[0.12\textwidth][c]{\textbf{Result}}%
        \makebox[0.12\textwidth][c]{\textbf{Seg}}%
    }
    \vbox{

    \centering
    \includegraphics[width=0.96\textwidth]{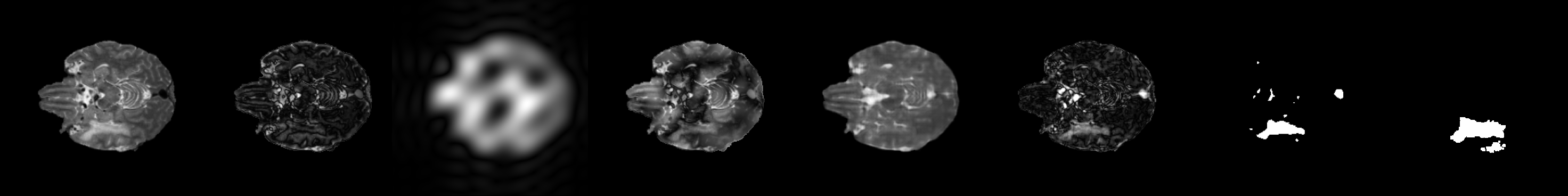}
    \vspace{-\baselineskip}

    \centering
    \includegraphics[width=0.96\textwidth]{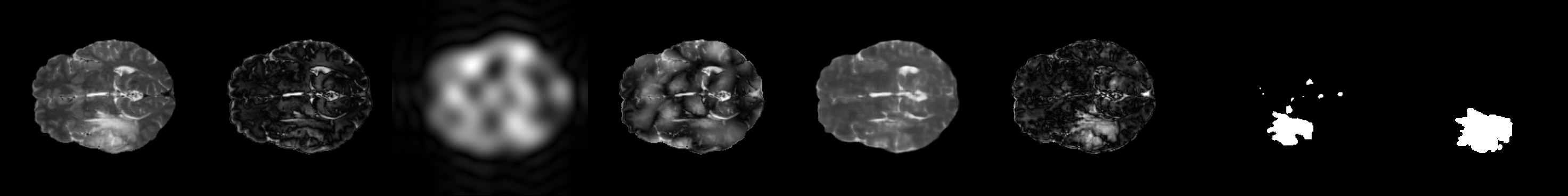}
    \vspace{-\baselineskip}

    \centering
    \includegraphics[width=0.96\textwidth]{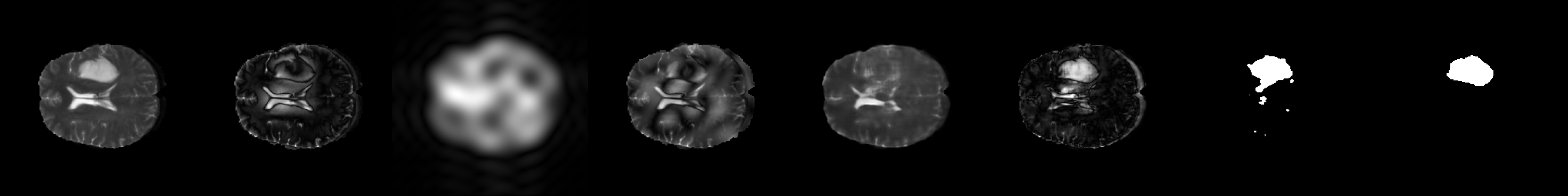}
    \vspace{-\baselineskip}

    \centering
    \includegraphics[width=0.96\textwidth]{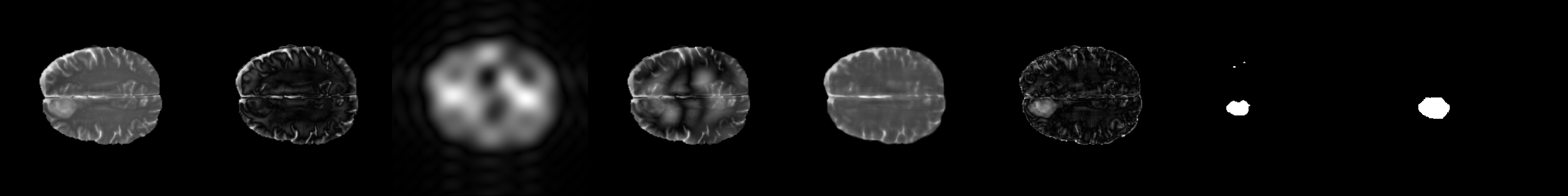}
    \vspace{-\baselineskip}
    
    \centering
    \includegraphics[width=0.96\textwidth]{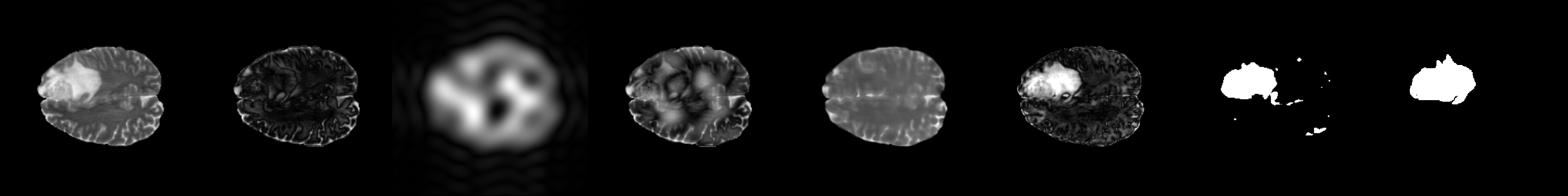}
    \vspace{-\baselineskip}
    
    \centering
    \includegraphics[width=0.96\textwidth]{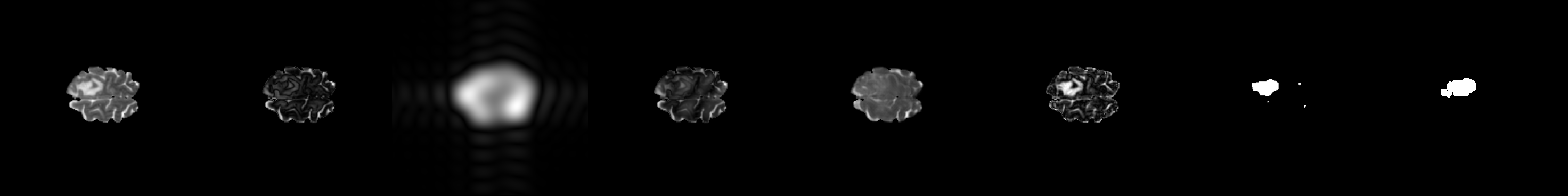}
    \vspace{-\baselineskip}
    
    }
    \vspace{1ex}
    \caption{The figure presents the intermediate and final results of MRI reconstruction. From left to right, the columns represent: the original image (Origin), the MRI with low-frequency components removed (RemoveLF), reconstructed low-frequency signals (LFRec), LFRec combined with original high-frequency signals (FRec), image generated by the model (Recon), residual map (Residual), final detection result (Result), and ground-truth lesion segmentation (Seg).}
    \label{fig:frq_rec}
\end{figure*}

\subsection{Visualization of MRI Reconstruction}
\label{Visualization_MRI_Reconstruction}
To more intuitively demonstrate the effectiveness of FDP in removing anomalies from MRI images, we present visualizations of results at each pipeline stage produced by FDP. 
As shown in Figure~\ref{fig:frq_rec}, the reconstructed low-frequency signals recover relatively healthy low-frequency information from the input images, preserving the structural characteristics of the MRI while attenuating the influence of lesion regions. At the same time, frequency reconstruction introduces some noise. Encouragingly, this noise can be readily learned by the subsequent generative model and effectively suppressed during inference.

\subsection{Visualization of Low-Frequency Prior Context}
We selected several low-frequency prior contexts and transformed them back into the spatial domain for visualization as images.

\begin{figure*}[th]
    \centering
    \includegraphics[width=0.55\linewidth]{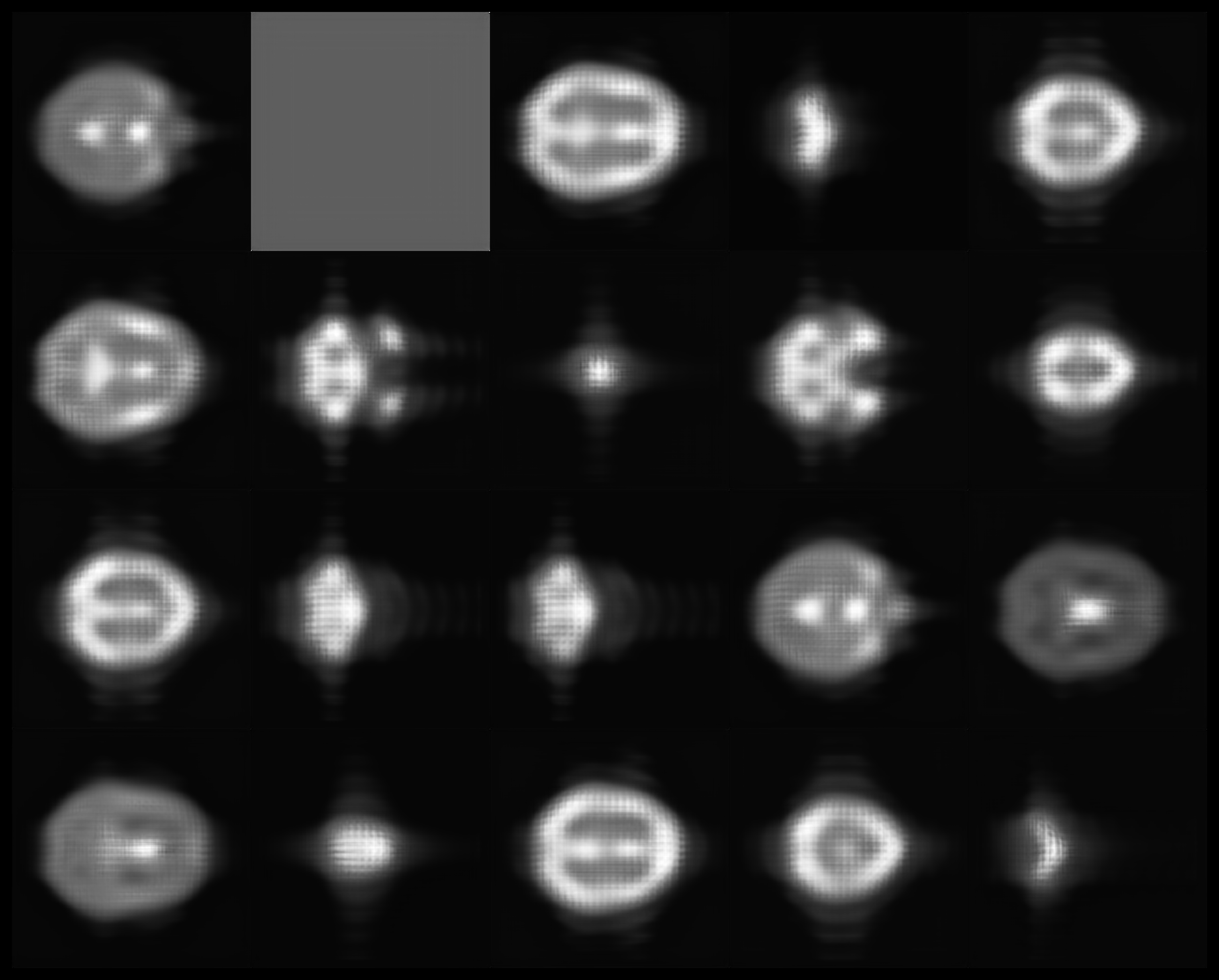}
    \caption{Visualization of Learnable Low-Frequency Prior Context.}
    \label{fig:low_freq_vis}
\end{figure*}

As shown in the Figure~\ref{fig:low_freq_vis}, the prior context encompasses a diverse range of MRI features, capturing variations in intensity and anatomical structure. Notably, the second prototype—an almost entirely empty image—is well-suited for handling cases where no brain tissue is present in the current MRI slice. These visualization results further support the effectiveness of our frequency prior context bank in modeling structural diversity across healthy brain scans.

\subsection{Evaluation Metrics}
\label{Evaluation_Metrics}
We used AUROC, AUPRC, and DICE as evaluation metrics for our method. For each slice of a volume, we computed the metrics over the effective area and then averaged the results across all slices to obtain the final performance scores. AUROC measures the model’s ability to distinguish between abnormal and normal samples, while AUPRC evaluates the trade-off between precision and recall across different thresholds. For anomaly detection, DICE is primarily used as a region-based segmentation metric to assess the similarity between the predicted and ground-truth lesion regions.

\subsection{Post-Processing}
\label{Post_Processing}
We used the reconstruction difference as the pixel-wise anomaly score $\delta = \|{I} - {I}_\mathrm{rec}\|$. To ensure consistency across MRI slices, we applied a 3D mean filter with a kernel size of $5$ before binarization. We then eroded the brain mask for three iterations. Higher anomaly scores correspond to larger reconstruction errors and therefore indicate a higher likelihood of abnormal regions. We determined the decision threshold by performing a greedy search on the unhealthy validation set, iteratively computing the DICE score for different thresholds. The optimal threshold was then used to compute the average DICE score on the unhealthy test set.

% \input{ReproducibilityChecklist}


\end{document}